\documentclass[letterpaper, 10 pt, conference]{ieeeconf}  

\IEEEoverridecommandlockouts                              

\usepackage{graphics} 
\usepackage{epsfig} 
\usepackage{mathptmx} 
\usepackage{times} 
\usepackage{amsmath} 
\usepackage{amssymb}  
\usepackage{xcolor}
\usepackage{bbding}
\usepackage{bbold}
\usepackage{hyperref}
\usepackage{booktabs}
\usepackage{graphicx}
\usepackage{subfigure}
\usepackage[noend]{algorithm2e}
\usepackage{algcompatible}%
\usepackage{amssymb}
\usepackage{pifont}
\newcommand{\cmark}{\ding{51}}%

\let\oldnl\nl
\newcommand{\nonl}{\renewcommand{\nl}{\let\nl\oldnl}}

\SetAlFnt{\small}

\definecolor{LQY_color}{RGB}{50, 205, 50}

\title{\LARGE \bf
Improving the Generalization of End-to-End Driving through \\ Procedural Generation
}

\author{Quanyi Li$^{*2}$, Zhenghao Peng$^{*1}$, Qihang Zhang$^{2,3}$, Chunxiao Liu$^{2}$, Bolei Zhou$^{1}$%
\thanks{$^{*}$ Quanyi Li and
Zhenghao Peng contribute equally to this work.}
\thanks{$^{1}$ Bolei Zhou and Zhenghao Peng are with the Department of Information Engineering, the Chinese University of Hong Kong, Shatin, N.T., Hong Kong
        {\tt\small \{bzhou, pengzh\}@ie.cuhk.edu.hk}}%
\thanks{$^{2}$ Chunxiao Liu, Quanyi Li and Qihang Zhang are with SenseTime Group Limited
        {\tt\small \{liuchunxiao, liquanyi1\}@sensetime.com}}%
\thanks{$^{3}$ Qihang Zhang is with the College of Computer Sciende and Technology, Zhejiang University, Hangzhou, China {\tt\small qh\_zhang@zju.edu.cn}}%
}

\begin{document}

\maketitle
\thispagestyle{empty}
\pagestyle{empty}

\begin{abstract}

Over the past few years there is a growing interest in the learning-based self driving system. To ensure safety, such systems are first developed and validated in simulators before being deployed in the real world. However, most of the existing driving simulators only contain a fixed set of scenes and a limited number of configurable settings. That might easily cause the overfitting issue for the learning-based driving systems as well as the lack of their generalization ability to unseen scenarios. To better evaluate and improve the generalization of end-to-end driving, we introduce an open-ended and highly configurable driving simulator called PGDrive, following a key feature of procedural generation. Diverse road networks are first generated by the proposed generation algorithm via sampling from elementary road blocks. Then they are turned into interactive training environments where traffic flows of nearby vehicles with realistic kinematics are rendered. We validate that training with the increasing number of procedurally generated scenes significantly improves the generalization of the agent across scenarios of different traffic densities and road networks. Many applications such as multi-agent traffic simulation and safe driving benchmark can be further built upon the simulator. To facilitate the joint research effort of end-to-end driving, we release the simulator and pretrained models at \url{https://decisionforce.github.io/pgdrive}.

\end{abstract}

\section{INTRODUCTION}

\begin{figure*}[!th]
    \centering
    \includegraphics[width=0.94\linewidth]{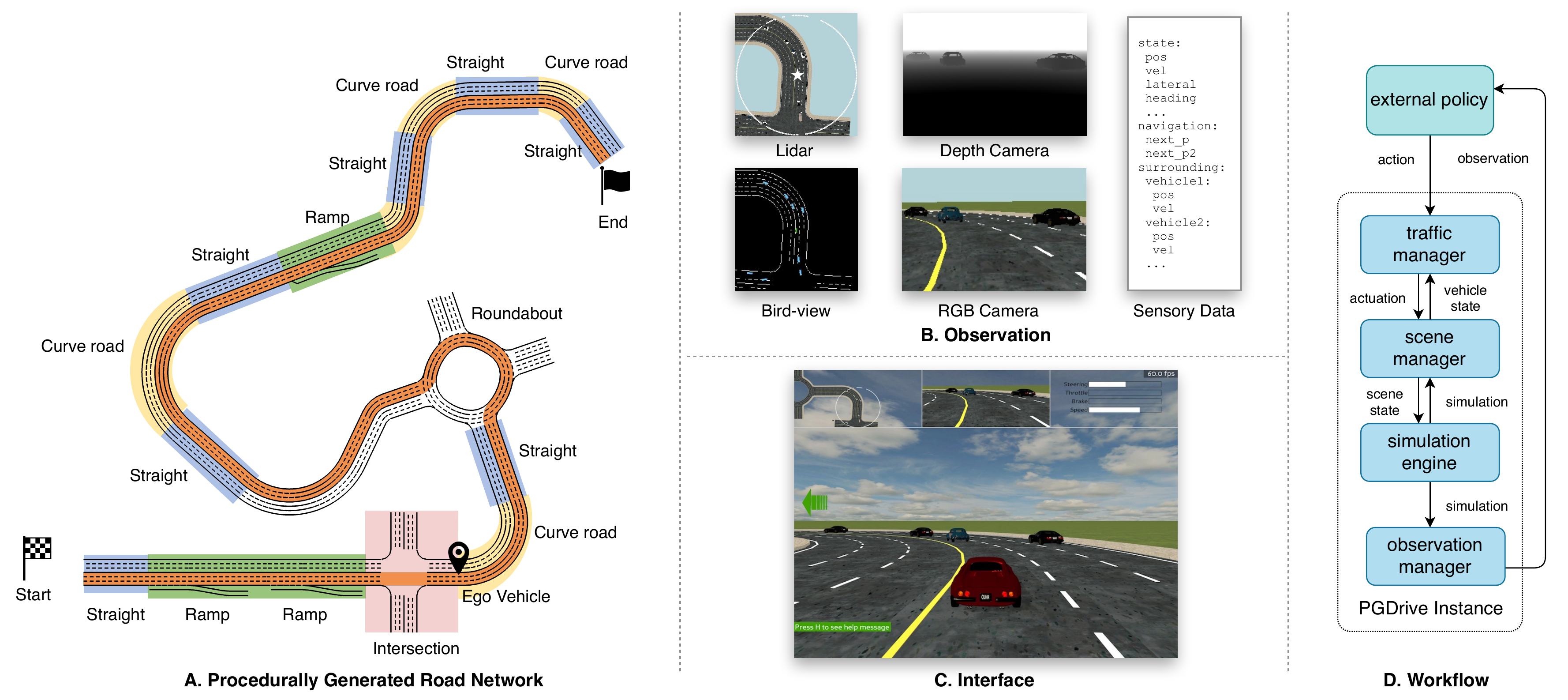}
    \caption{\textbf{A.} A procedurally generated map built from elementary road blocks. \textbf{B.} Multi-modal observations provided by PGDrive, including Lidar-like cloud points, RGB / depth camera, bird-view semantic map and scalar sensory data. \textbf{C.} Interface of the PGDrive simulator for visualization and manual control. \textbf{D.} Simulation pipeline in run-time.}
    \label{fig:teaser}
\end{figure*}

In recent years there has been a rapid progress in self driving in both industry and academia~\cite{Montemerlo2008Junior,yurtsever2020survey}.
Self driving pipeline can be split into two categories, the modular system and the end-to-end learning-based system. 
The modular system divides the driving into multiple modules such as perception, decision, planning, and control where each module need to be tuned carefully to maximize the joint performance.
The end-to-end approach for self driving, where the entire driving pipeline is replaced by a single deep neural network, need to be trained with a massive number of data or interactions with the environment, for instance, in Imitation Learning~\cite{DBLP:journals/corr/BojarskiTDFFGJM16} or Reinforcement Learning~\cite{Riedmiller2007Learning}.
A dilemma in prototyping aforementioned systems is that, due to public safety concerns, it is impossible to directly develop, train, and test the self driving systems in the physical world~\cite{tampuu2020survey}.
To overcome the dilemma, a common practice is to prototype and validate the driving system in simulators.
Many realistic driving simulators have been proposed. 
CARLA~\cite{Dosovitskiy17} and SUMMIT~\cite{cai2020summit} contain driving scenarios with realistic visual appearances. Particularly, CARLA enables the fine-grained rendering of the real world with realistic weather and lighting conditions. However, the efficiency of CARLA simulator is not ideal, with only about 25 simulation steps per second on a single instance in PC.
Other simulators like Flow~\cite{vinitsky2018benchmarks}, Duckietown~\cite{gym_duckietown}, and Highway-env \cite{highway-env} either focus on learning high-level decisions or only contain simplistic driving scenarios. 
However, it's notorious that neural networks can overfit training data easily~\cite{zhang2018study,cobbe2019quantifying}, thus the driving agent trained on a fixed set of scenarios in the simulators might have difficulty generalizing to new scenarios. 
Furthermore, the existing simulators lack sufficiently diverse designs of the maps and scenarios to evaluate such issues. For example, CARLA consists of only about ten manually designed towns, which is hard to modify, while Highway-env contains six typical traffic scenarios. 
Training deep neural networks on such simulators might lead to overfitting and poor generalization of learned driving policy.

To evaluate and improve the generalization of end-to-end driving, we introduce PGDrive, an open-ended and highly configurable driving simulator. PGDrive integrates the key feature of the procedural generation (PG) \cite{shaker2016procedural,risi2020increasing}, where a diverse range of executable driving scenes can be procedurally generated.
Such scenes can be used to train and test driving systems for examining their generalization ability. 
Specifically, we abstract the road network as a combination of several road blocks, such as Straight, Ramp, Fork, Roundabout, Curve, and Intersection, with configurable settings. Following the proposed PG algorithm, these road blocks can be randomly selected and assembled into an interactive environment to drive in, as illustrated in Fig.~\ref{fig:teaser}A.
Apart from the road network, the kinematics of target vehicle and traffic vehicles, the multi-modal observations (Fig.~\ref{fig:teaser}B) are also configurable for emulating various traffic flows, transportation conditions, and input data flow.
Thus the key feature of procedural generation in PGDrive allows us to customize the training set, examine different emergent driving behaviors, and improve the model's generalization across scenarios. 
PGDrive is built upon Panda3D and the Bullet engine \cite{goslin2004panda3d} with optimized system design. It can achieve up to 500 simulation steps per second when running a single instance on PC, and can be easily paralleled to further boost efficiency.

Based on the proposed PGDrive simulator, we study the generalization of learning-based driving systems. We first validate that the agent trained on a small fixed set of maps generalizes poorly to new scenarios. The experimental result further shows that training with more environments, the learned reinforcement learning agents can generalize better to unseen scenarios in terms of less collision rate and traffic violation. It demonstrates the benefit of procedural generations brought by PGDrive. 
We also raise a concerning issue called safety generalization. We find that the overfitting of safety performance also exists, and the policy trained with a limited number of scenes yields frequent crashes.
We open-source PGDrive to facilitate the research of end-to-end driving.

\section{RELATED WORK}

\textbf{Self Driving.}
The research on self driving can be divided into two categories: the modular and the end-to-end approaches~\cite{yurtsever2020survey}.
In the modular approach, the driving pipeline is composed of sub-modules such as perception, localization, planning, and control module~\cite{yurtsever2020survey}. The advantage of the modular design is the interpretability, where one can easily identify the module at fault in case of a malfunction or unexpected system behavior. However, it takes enormous human efforts to design and maintain the pipeline, and it might also suffer from the internal redundancy caused by manual designs \cite{tampuu2020survey}.
On the other hand, one popular end-to-end approach is to utilize the Imitation Learning (IL) to drive a car by mimicking an expert \cite{tampuu2020survey}. The accessibility of large amounts of human driving data \cite{bojarski2016end} makes the IL approach work well for simple tasks such as lane following \cite{pomerleau1989alvinn} and obstacle avoidance \cite{muller2006off}. Nevertheless, the training data might be biased towards particular locations, road layouts, or weathers \cite{codevilla2019exploring}, while the accident cases can rarely occur in the training data but deserve significant attention \cite{tampuu2020survey}. In contrast to the aforementioned approaches, Reinforcement Learning (RL) gains attention recently as another approach of end-to-end driving. RL is able to discover task-specific knowledge by interaction with the environment~\cite{sutton2018reinforcement}, thus the needs for elaborate design and domain knowledge are greatly alleviated. Meanwhile, RL agent learns through the exploration of the environment thus does not limit its capacity to the training data. Based on the strengths, more and more attention has been paid to the RL-based solutions \cite{kiran2020deep}.

\textbf{Driving Simulators.} 
Due to public safety and cost concerns, it is impossible to train and test the driving models in the physical world on a large scale. Therefore, simulators have been used extensively to prototype and validate the self driving research.
The simulators CARLA~\cite{Dosovitskiy17}, GTA V~\cite{martinez2017beyond}, and SUMMIT~\cite{cai2020summit} realistically preserve the appearance of the physical world. 
For example, CARLA not only provides perception, localization, planning, control modules, and sophisticated kinematics models, but also renders the environment in different lighting conditions, weather and the time of a day shift. Thus the driving agent can be trained and evaluated more thoroughly.
Other simulators such as Flow \cite{vinitsky2018benchmarks}, TORCS \cite{wymann2000torcs}, Duckietown~\cite{gym_duckietown} and Highway-env \cite{highway-env} abstract the driving problem to a higher level or provide simplistic scenarios as the environments for the agent to interact with. SMARTS~\cite{zhou2020smarts} provides playground for learning multi-agent interaction in self driving.
However, most of the existing simulators have a limited number of maps thus fail to provide the diverse training environments for the agent to learn and generalize. It remains challenging to evaluate the generalization of the trained agents. Moreover, it's hard to modify the existing simulator to apply the procedural generation technique we used in this work. For example, in CARLA, the town is manually designed with hard-coded buildings, road structures and so on. To address the generalization issue, we abstract the driving scene and integrate the procedural generation (PG) in the proposed simulator to generate a huge amount of diverse training and test data.

\textbf{Procedural Generation.}
Procedural Generation (PG) or Procedural Content Generation (PCG) are first adopted in the video game industry~\cite{shaker2016procedural}. It refers to the practice of utilizing algorithms to automatically generate game content including levels, maps, racing tracks, etc~\cite{herrera1998automatic, georgiou2016personalised}. Many machine learning concepts like data augmentation~\cite{krizhevsky2012imagenet}, domain randomization~\cite{ tobin2017domain} and Generative Adversarial Network~\cite{goodfellow2014generative} have been proposed to help design PG algorithms, making the game playable and alleviating the burden of game designers. 
The machine learning community also becomes interested in applying PG to improve the learning models~\cite{risi2020increasing}.
Researchers have utilized PG to generate abundant training samples along with various training settings to help relieve data over-fitting, enable lifelong learning with lifelong generation~\cite{thrun1995lifelong, parisi2019continual}, and help agents adapt to real-world from learning in simulated settings~\cite{kar2019meta}.

\textbf{Generalization.}
Attaining generalization from training has long been a focus of machine learning research \cite{schmidhuber2015deep,allen2019learning}. Models with weak generalization usually perform poorly with unseen data, despite the excellent performance on the training dataset~\cite{hardt2016equality, lin2016generalization}.
In reinforcement learning (RL), overfitting exists but is usually overlooked. In most RL settings such as Deep-Q networks on Atari games \cite{mnih2015human}, the training and testing are conducted in the same environment and thus the agents are prone to memorizing specific features in the environment~\cite{zhang2018study}. Nichol et al.~\cite{nichol2018gotta} propose to measure generalization performance by training and testing RL agents on distinct sets of levels in a video game franchise, which reveals the limitation of \textit{training on the test set} paradigm. Cobbe et al.~\cite{cobbe2019quantifying} propose to use PCG to generate many distinct levels of video games and split the training and test sets to quantify the generalization of policy. 

\section{PGDrive Simulator}

To benchmark the generalization ability of a driving policy, we introduce a new driving simulator PGDrive, which can generate an unlimited number of diverse driving environments through procedural generation. 
As depicted in Table~\ref{tab:comparison}, PGDrive preserves unique features compared to other popular driving simulators.
The key feature of PGDrive is Procedural Generation (PG). Unlimited number of diverse driving scenes can be generated with the help of our abstraction on driving scene as well as the proposed PG algorithm.
PGDrive is highly configurable, allowing users to customize settings such as the traffic flow and the vehicles' dynamics models. 
Besides, fine-grained camera, Lidar data generation, and realistic 3D kinematics simulation are also supported. 
To support fast prototyping of driving system, PGDrive is designed to be lightweight and highly efficient, compared to other simulators which are expensive to run or difficult to install. Single PGDrive process can run up to 500 simulation steps per second without rendering on a PC and can be easily paralleled to further boost efficiency.

The key function of PGDrive is to generate unlimited number of diverse driving scene. We decompose a driving scene into several components: (1) the static \textit{road network}, which consists a set of \textit{road blocks} and the interconnection between them, and (2) the \textit{traffic flow}, containing a set of \textit{traffic vehicles} and a \textit{target vehicle}. 
In Sec.~\ref{sect:abstraction-of-scene}, we first describe the details of the abstracted road blocks. Then in Sec.~\ref{sect:pg-algorithm}, we propose a PG algorithm that combines those blocks to a road network. In Sec.~\ref{sect:RL-env}, we describe the run-time workflow of PGDrive. Other implementation details can be found in Sec.~\ref{sect:implementation-details}.

\begin{table}[!t]
\caption{Comparison of representative driving simulators}
\label{tab:comparison}
\resizebox{\linewidth}{!}{%
\begin{tabular}{ccccccc}
\toprule
Simulator & \begin{tabular}[c]{@{}c@{}}Unlimited\\ Maps\end{tabular} & \begin{tabular}[c]{@{}c@{}}Traffic\\ Density\end{tabular} & \begin{tabular}[c]{@{}c@{}}Custom\\ Vehicle\end{tabular} & \begin{tabular}[c]{@{}c@{}}Lidar or\\ Camera\end{tabular} & \begin{tabular}[c]{@{}c@{}}Realistic\\Kinematics\end{tabular} \\ \midrule\midrule
CARLA~\cite{Dosovitskiy17}           & &  &\cmark &\cmark &\cmark  \\ \midrule 
GTA V~\cite{martinez2017beyond}           & &  &\cmark &\cmark &\cmark\\ \midrule
Highway-env~\cite{highway-env}         & &\cmark  & & & \\ \midrule 
TORCS~\cite{wymann2000torcs}           & & & &\cmark &\cmark \\ \midrule
Flow~\cite{vinitsky2018benchmarks}           & &\cmark  & & &\cmark\\\midrule 
Sim4CV~\cite{muller2018sim4cv}          & &  & &\cmark & \\\midrule 
Duckietown~\cite{gym_duckietown} & & \cmark & & \cmark & \\ \midrule %
SMARTS~\cite{zhou2020smarts} & & \cmark & & \cmark & \cmark \\ \midrule %
\begin{tabular}[c]{@{}c@{}}\textbf{PGDrive}\\\textbf{(Ours)}\end{tabular}   &\cmark &\cmark  &\cmark &\cmark & \cmark\\ \bottomrule
\end{tabular}
}
\end{table}

\begin{figure*}[!t]
\centering
\includegraphics[width=0.96\linewidth]{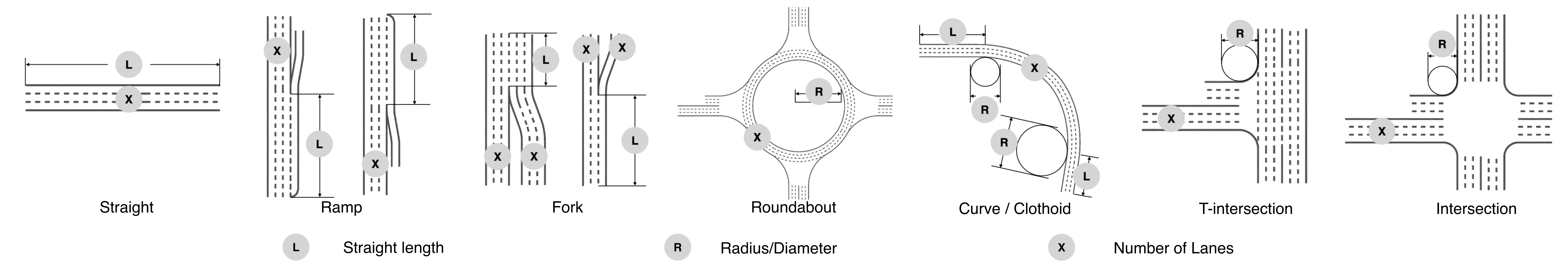}
\caption{Seven types of road blocks and their parameters. \textbf{L}, \textbf{R}, \textbf{X} indicate the road length, the road curvature and the number of lanes, respectively.
}
\label{fig:block_type}
\end{figure*}

\subsection{Road Blocks}
\label{sect:abstraction-of-scene}

\RestyleAlgo{ruled}
\begin{algorithm}[!t]
\caption{Procedural Generation of Driving Scenes}
\label{algo:pg}
\SetAlgoLined
\LinesNumbered
\DontPrintSemicolon
\KwIn{
Maximum tries in one block {\tt T}; Number of blocks in each map {\tt n}; Number of required maps {\tt N}\\
}
\KwResult{
 A set of maps $M =\{G^{(i)}_{net}\}_{i=1, ..., N}$
}
\SetKwProg{Fn}{Function}{}{}
\# Define the main function to generate a list of maps~\\
\Fn{{\tt \textbf{main(}T, n\textbf{)}}}{
    Initialize an empty list $M$ to store maps \\
     \While{$M$ does not contain {\tt N} maps}{
        Initialize an empty road network $G_{net}$ \\
        $G_{net}$, {\tt success}=\textbf{\textit{BIG({\tt T}, $G_{net}$, {\tt n})}} \\
        \uIf{{\tt success} is {\tt True}}{
        	Initialize traffic vehicles in some spawn points $G_{net}(P)$ \\
            Append $G_{net}$ to $M$
        }
    }
\textbf{Return} {\tt M}
}
\nonl ~\\
\# Define the Block Incremental Generation (BIG) helper function that appends one block to current map if feasible and return current map with a success flag ~\\
\Fn{{\textbf{BIG(}{\tt T}, $G_{net}$, {\tt n}\textbf{)}}}{
    \uIf{$G_{net}$ has {\tt n} blocks}{
         \textbf{Return} {$G_{net}$, {\tt True}} 
    }%
    \For{{\tt 1, ..., T}}{
        Create new block $G_{\omega}$ = \textbf{\textit{GetNewBlock()}} \\
        Find the sockets for new block and old blocks: $e_1 \sim G_{\omega}(S)$, $e_2 \sim G_{net}(S)$ \\
        Rotate $G_{\omega}$ so that $e_1$, $e_2$ have supplementary heading\\
        \uIf{$G_{\omega}$ does not intersect with $G_{net}$}{
             $G_{net} \gets G_{net} \bigcup G_{\omega}$ \\
             $G_{net}$, {\tt success}=\textbf{\textit{BIG({\tt T},$G_{net}$, {\tt n})}} \\
            \uIf{{\tt success} is {\tt True}}{
                 {\textbf{Return} { $G_{net}$, {True}}}
            }
            \uElse{
                Remove the last block from $G_{net}$
            }
        }
    }
    \textbf{Return} $G_{net}$, {\tt False}
}
\nonl ~\\
\# Randomly create a block ~\\
\Fn{{\textbf{GetNewBlock()}}}{
    Randomly choose a road block type $T \sim \{ \text{Staright}, ... \}$ \\
    Instantiate a block and randomize the parameters $G_{\omega}, \omega \sim \Omega_T$ \\
    \textbf{Return} $G_{\omega}$
}
\end{algorithm}

As shown in Fig.~\ref{fig:block_type}, we define seven typical types of road block. We represent a road block using an undirected graph with additional features: $G_\omega=\{V, E, S, P, \omega, \Omega\}$, with nodes $V$ denoting the joints in the road network and edges $E$ denoting \textit{lanes} which interconnects nodes. At least one node is assigned as the \textit{socket} $S=\{e_{i_1}, e_{i_2}, ...\}, e \in E$. 
The socket is usually a straight road at the end of lanes which serves as the anchor to connect to the socket of other block. 
Block can preserve several sockets. For instance, Roundabout and T-Intersection have 4 sockets and 3 sockets, respectively.
There are also some \textit{spawn points} $P$ distributed uniformly in the lanes for allocation of traffic vehicles. Apart from the above properties, a block type-specific \textit{parameter space} $\Omega_{T}$ is defined to bound the possible parameters $\omega$ of the block, such as the number of lanes, the lane width, and the road curvature and so on. $T$ is the block type from 7 predefined types.
 The road block is the elementary traffic component that can be assembled into a complete road network $G_{\omega_{net}} = \{ V, E, S, P, \omega_{net} \} = G_{\omega_1} \bigcup ... \bigcup G_{\omega_n}$ by the procedural generation algorithm. 
Shown in Fig.~\ref{fig:block_type}, the detail of each block type $T$ is summarized as follows:
\begin{itemize}
    \item \textbf{Straight:} A straight road is configured by the lanes number, length, width, and types, namely whether it is broken white line or solid white line.
    \item \textbf{Ramp:} A ramp is a road with entry or exit existing in the rightest lane. Acceleration lane and deceleration lane are attached to the main road to guide the traffic vehicles to their destination.
    \item \textbf{Fork:} A structure used to merge or split additional lanes from the main road and change the number of lanes.
    \item \textbf{Roundabout:} A circular junction with four exits (sockets) with configurable radius. Both roundabout, ramp and fork aim to provide diverse merge scenarios.
    \item \textbf{Curve:} A curve block consists of circular shape or clothoid shape lanes with configurable curvature.
    \item \textbf{T-Intersection:} An intersection that can enter and exit in three ways and thus has three sockets. The turning radius is configurable.
    \item \textbf{Intersection:} A four-way intersection allows bi-directional traffic. It is designed to support the research of unprotected-intersection.
\end{itemize}

\subsection{Procedural Generation of Driving Scene}
\label{sect:pg-algorithm}
After defining the road blocks, we use the Procedural Generation (PG) technique to automatically select and assemble these blocks into diverse driving scenes. 
As illustrated in Algorithm~\ref{algo:pg}, we propose a search-based PG algorithm \textit{Block Incremental Generation (BIG)}, which recursively appends block to the existing road network if feasible and reverts last block otherwise. 
When adding new block, BIG first uniformly chooses a road block type $T$ and instantiate a block $G_{\omega}$ with random parameters $\omega\sim \Omega_T$ (the function \textbf{\textit{GetNewBlock()}}). After rotating the new block so the new block's socket can dock into one socket of existing network (Line 17, 18), BIG will then verify whether the new block intersects with existing blocks (Line 19). We test the crossover of all edges of $G_{\omega}$ and network $G_{net}$. If crossovers exist, then we discard $G_{\omega}$ and try new one. Maximally {\tt T} trials will be conducted. If all of them fail, we remove the latest block and revert to the previous road network (Line 25).

We set the stop criterion of BIG to the number of blocks. After generating a road network with {\tt n} blocks, the initial traffic flow is attached to the static road network (Line 8) to complete the scene generation by the \textit{traffic manager}.
The traffic manager creates traffic vehicles by randomly selecting the vehicle type, kinematics coefficients, target speed, behavior (aggressive or conservative), spawn points and destinations. 
The traffic density is define as the number of traffic vehicles per lane per 10 meters and is considered in this period. We randomly select $D\times L \times X / 10$ spawn points in the whole map to allocate the traffic vehicles, wherein $D$ is the given traffic density, $L$ is the total length of road network, and $X$ is the average number of lanes. 
In Sec.~\ref{sect:factors-relevant-to-generalization}, we show the learning system can also overfit to the traffic density, which highlights the importance of varying traffic conditions during training.
We introduce the generation process of scene in this section. In the following, we will introduce the run-time workflow of PGDrive.

\subsection{Run-time Workflow}
\label{sect:RL-env}

As shown in Fig.~\ref{fig:teaser}D, a PGDrive instance contains a traffic manager, a scene manager, a simulation engine and an observation manager. 
After initialization, at each time step, the policy will pass the normalized control signals of the target vehicle to PGDrive. The traffic manager converts the normalized action $\mathbf{a}_{ext} = [a_1, a_2]^T \in [-1, 1]^{2}$ into the steering $u_{s}$, acceleration $u_{a}$ and brake signal $u_{b}$ in the following ways:
\begin{eqnarray}
	u_{s} = & a_1 \times 40 &\text{(degree)}, \\
	u_{a} = & \max(0, a_2) \times 460 &\text{(hp)}, \\
	u_{b} = & -\min(0, a_2) \times 355 &\text{(hp)}.
\end{eqnarray}

Traffic manager will also query the IDMs for the actuation of traffic vehicles.
All actuation signals are then passed to the scene manager, augmented by other scene states such as light conditions and camera positions. The scene manager communicates with the simulation engine and advances the scene for $0.1 s$ interval. The observation manager retrieves information from simulation engine and scene manager and provides various forms of observation, as shown in Fig.~\ref{fig:teaser}B.

\begin{figure}[!t]
    \centering
   	\includegraphics[width=0.95\linewidth]{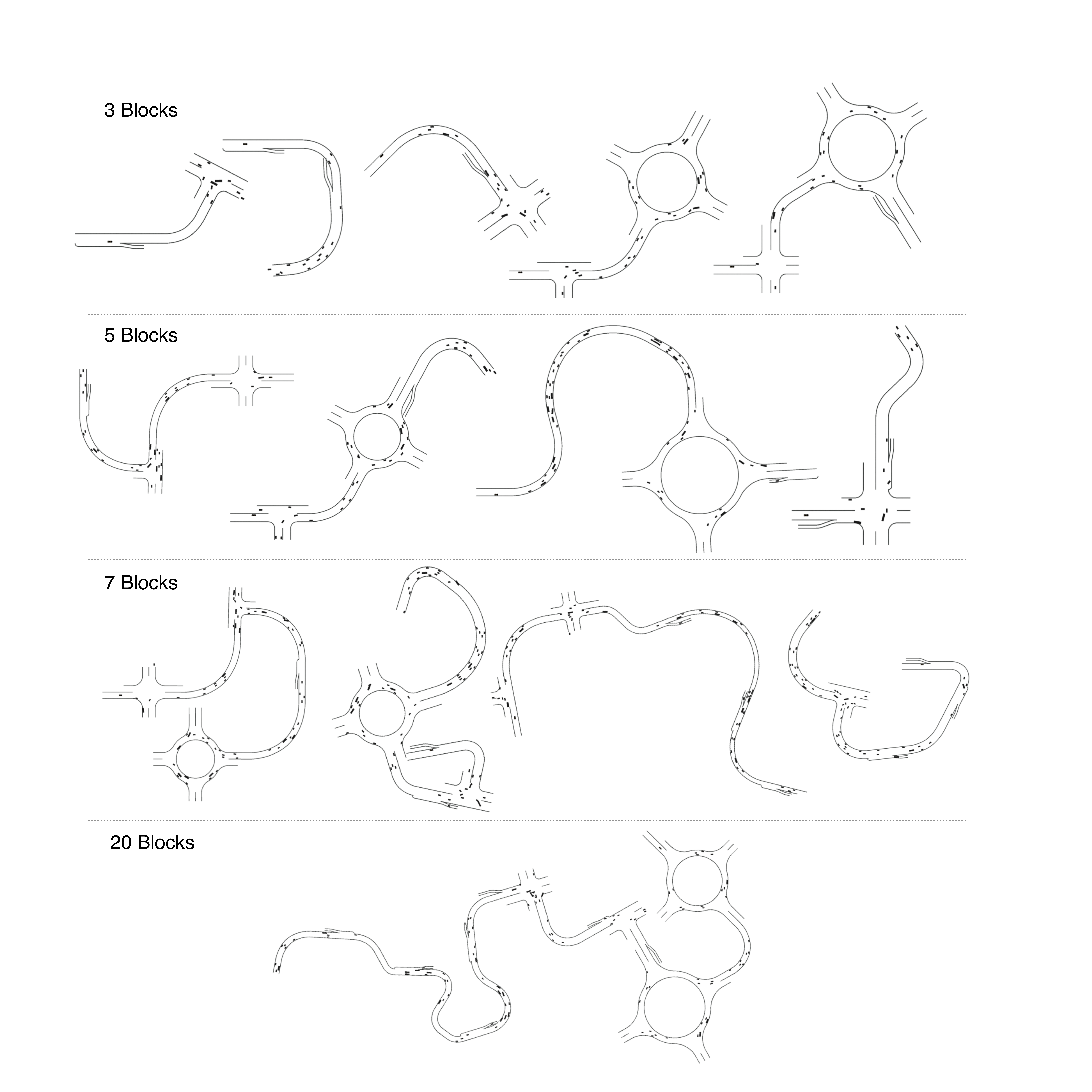}
    \caption{Procedurally generated scenes with different number of road blocks.}
    \label{fig:maps}
\end{figure}

\subsection{Implementation Details}
\label{sect:implementation-details}

\textbf{Simulation Engine.} PGDrive is implemented based on Panda3D~\cite{goslin2004panda3d} and Bullet Engine.
Panda3D is an open-source engine for real-time 3D games, rendering, and simulation. Its well designed rendering capacity enables PGDrive to construct realistic monitoring and observational data.
Bullet Engine is a physics simulation system that supports advanced collision detection, rigid body motion, and kinematic character controller, which empowers accurate and efficient physics simulation in PGDrive.

\textbf{Physical and Rendering Model.}
To improve the simulation efficiency and fidelity, all the elements in PGDrive, such as the lane lines, vehicles and even the Lidar lasers, are constructed by two internal models: the \textit{physical model} and the \textit{rendering model}.
Powered by Bullet engine, the physical model is a box shape rigid body that can participate in the collision detection and motion calculation. For instance, the collision between the physical models of the vehicle's wheels and the solid white line in the ground indicates the vehicle is driving out of road. The collision of Lidar lasers and rigid body can be use to compose the observation of Lidar-like cloud points.
The physical model contains rich configurable parameters, such as the wheel friction, max suspension travel distance, suspension stiffness, wheel damping rate and so on, supporting the customization of a vast range of vehicles in the environments created by PGDrive.
On the other hand, the \textit{rendering model}, which is loaded to Panda3D's render module, provides fine-grained rendering effects such as light reflection, texturing and shading.

\begin{figure}[!t]
    \centering
    \hfill%
    \subfigure[Multi-agent Environment]{
        \includegraphics[width=0.45\linewidth]{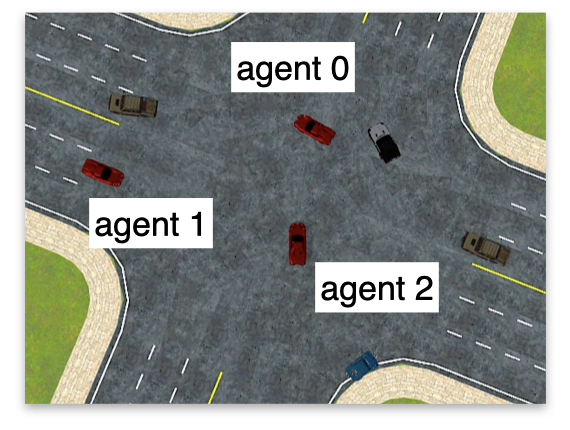}
    }    \hfill%
    \subfigure[Safety Environment]{
        \includegraphics[width=0.45\linewidth]{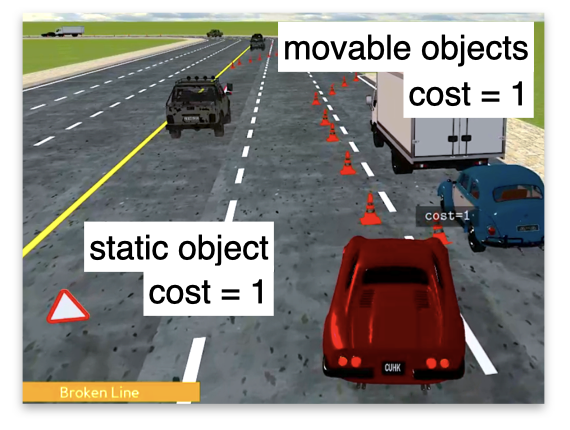}
    }    \hfill%
    \caption{(a) The multi-agent unprotected intersection scenario. (b) The safe driving environments with static and movable objects that mimic the dangerous situations.
}
\label{fig:applications}
\end{figure}

\begin{figure*}[!t]
    \centering    
    \begin{minipage}{0.35\linewidth}
    \includegraphics[width=\linewidth]{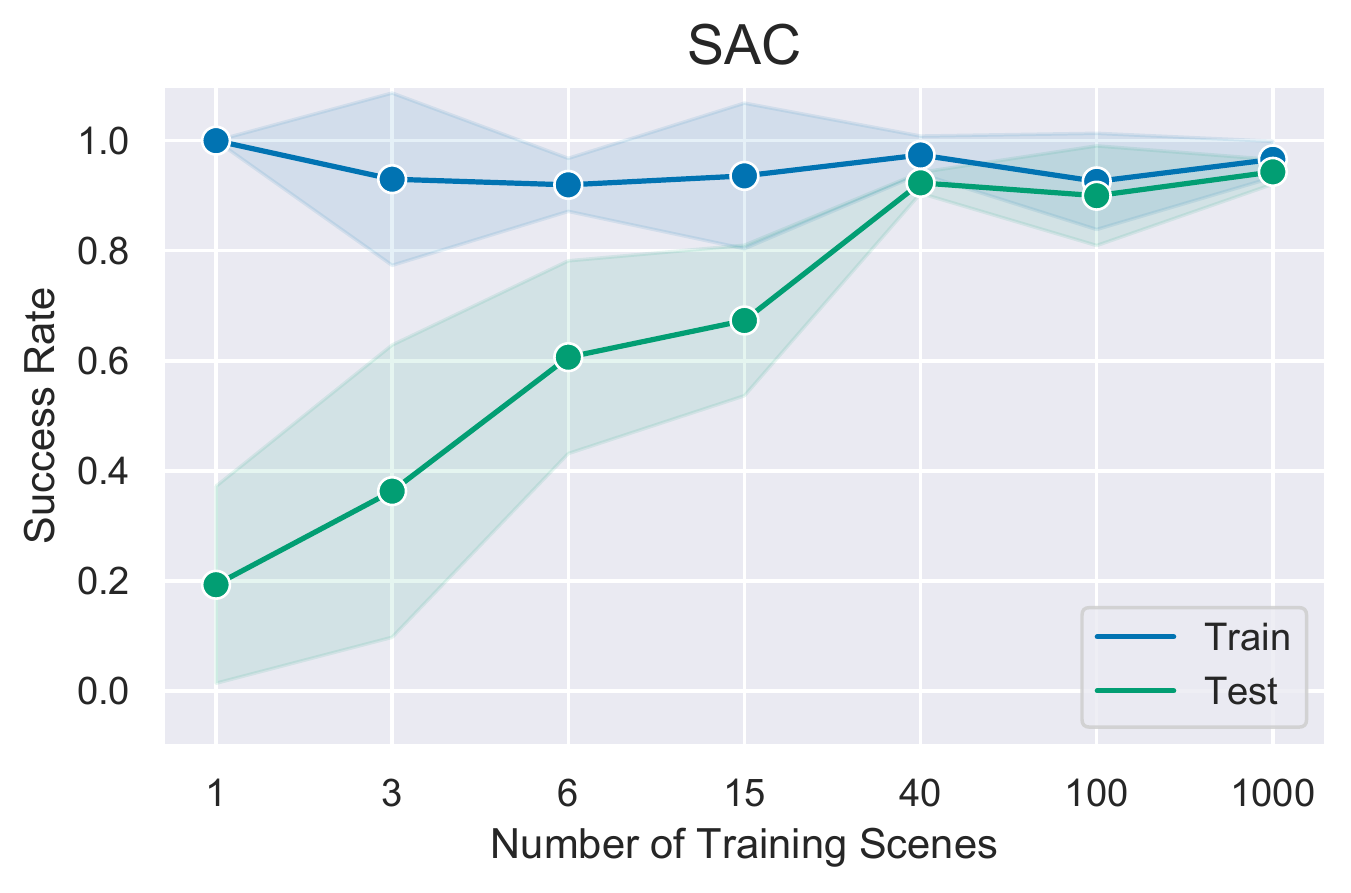}
    \end{minipage}\hfill%
    \begin{minipage}{0.35\linewidth}
    \includegraphics[width=\linewidth]{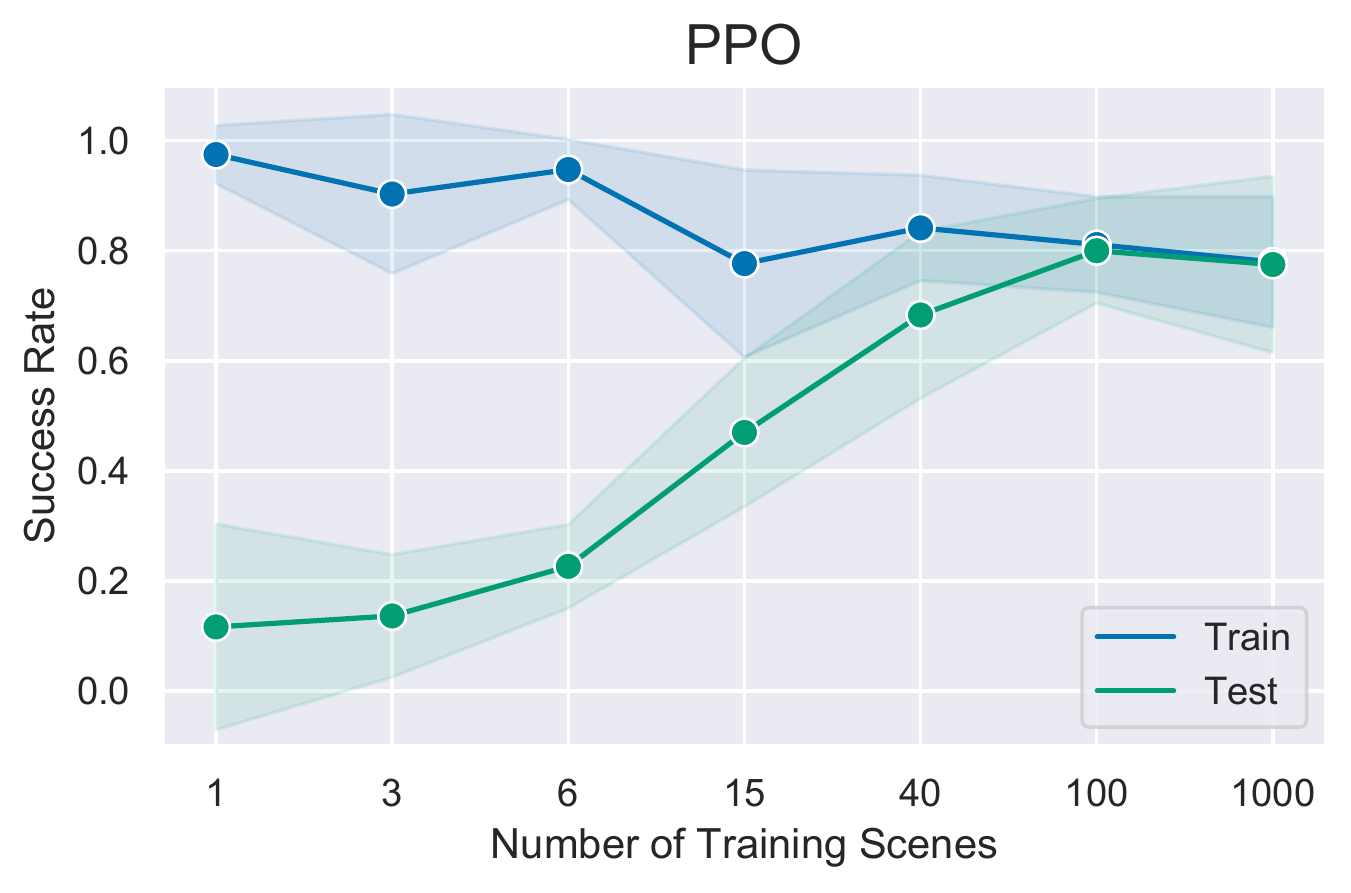}
    \end{minipage}%
    \hfill%
    \begin{minipage}{0.24\linewidth}
    \includegraphics[width=\linewidth]{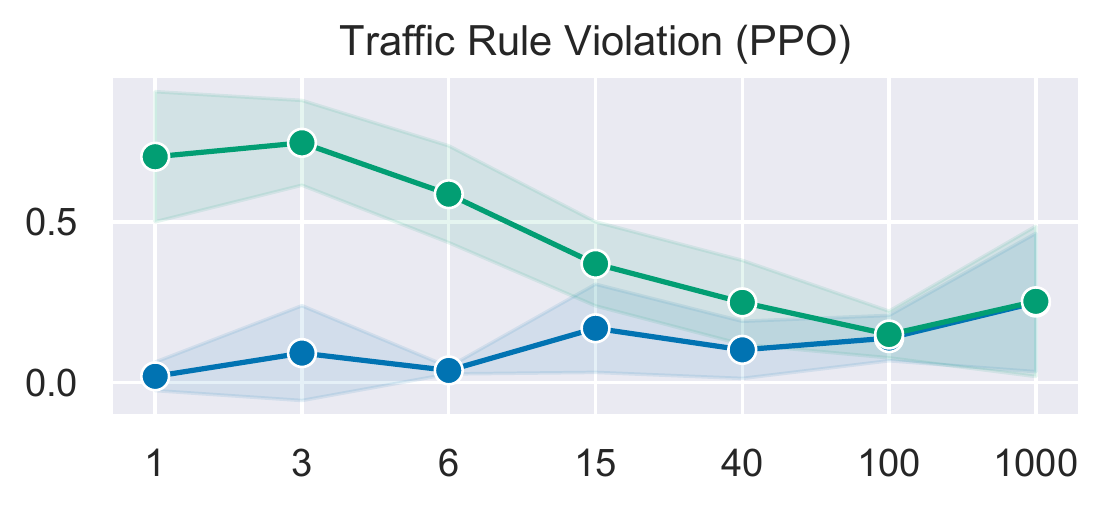}
    \includegraphics[width=\linewidth]{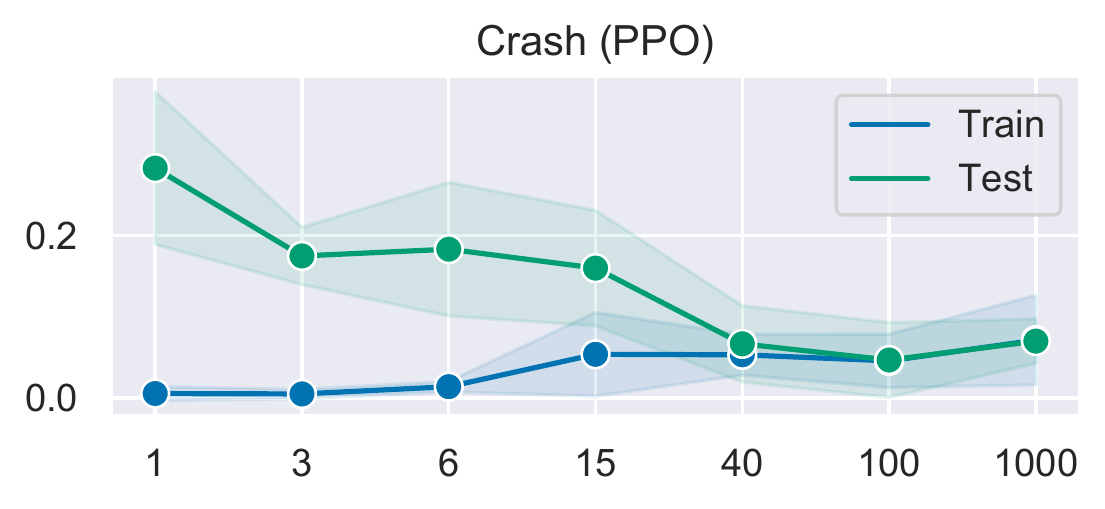}
    \end{minipage}%
    \hfill%
   	\caption{The generalization result of the agents trained with off-policy RL algorithm Soft Actor-critic (SAC)~\cite{haarnoja2018soft} and on-policy RL algorithm PPO~\cite{schulman2017proximal}.
    Increasing the number of training scenes leads to higher test success rate and lower traffic rule violation and crash probability, which indicates the agent's generalization is significantly improved.  
    Compared to PPO, SAC algorithm brings more stable training performance. The shadow of the curves indicates the standard deviation. 
    }
    \label{fig:main-result}
\end{figure*}

\textbf{Perception. } PGDrive provides various optional forms of observation to feed the external policy, as illustrated in Fig.~\ref{fig:teaser}B.
For low-level sensors, an unlimited number of RGB cameras, depth cameras and Lidar can be added with adjustable parameters, such as field of view, size of film, focal length, aspect ratio and the number of lasers.
Meanwhile, the high-level observation, namely the abstraction of surroundings, including the road information such as the bending angle, length and direction, and opponent vehicles' information like velocity, heading and profile, can also be provided as input directly without perception noise, thus enables the development of driving pipeline where the perception is assumed to be solved.

\textbf{Traffic Vehicle Policy. } A large number of traffic vehicles cruise in the scene from their spawn points to the randomly assigned destinations. The cruise behavior is powered by the IDM (Intelligent Driving model) with preset target speed, a parameter that randomized by the traffic manager. Apart from cruising, if the traffic vehicle is too closed to the front vehicle, the emergent stopping will be conducted by a pre-defined heuristic based on IDM. Furthermore, traffic vehicles can determine whether to change lane with a well-defined bicycle model-based lane change policy.

\textbf{Benchmarks and Applications.} Many driving applications can be built upon the proposed PGDrive simulator. The first is the generalization benchmark. As shown in Fig.~\ref{fig:maps}, different sets of scenes can be generated. So we can easily evaluate the driving systems' generalization ability in held-out set of scenes. Meanwhile, as shown in Fig.~\ref{fig:applications}~(a), it is easy to implement the scenarios of multi-agent reinforcement learning, since all vehicles in PGDrive share the same underlying structure so that we can assign arbitrary vehicles as the target vehicles of external policies. Besides, we can include obstacles in the traffic scene and model the collision as a soft cost for the driving agent as shown in Fig.~\ref{fig:applications}~(b), thus we can study the constrained optimization or safe exploration problem in reinforcement learning similar to the safety Gym~\cite{safety_gym_Ray2019}.
In the next experiment section, we will mainly demonstrate that the procedural generation in PGDrive benchmarks improves the generalization ability of end-to-end driving. 

\section{EXPERIMENTS}

Based on the proposed PGDrive simulator, we first conduct experiments to reveal the overfitting issue in the end-to-end driving agent, and show the improvement of the generalization brought by the PG technique. 
Furthermore, for the first time, we reveal the issue of safety generalization. We find that the overfitting to the training environment, in term of the safety performance, is also prevalent, even for the algorithm specialized at solving the constrained MDP.

\subsection{Experimental Setting}
\label{sect:exp-setting}

\textbf{Observation.}
We use the following information as the observation of RL agents:
\begin{itemize}
    \item A vector of length of 240 denoting the Lidar-like cloud points with $50 m$ maximum detecting distance.
    \item A vector containing the data that summarizes the target vehicle's state such as the steering, heading, velocity and relative distance to the left and right boundaries.
    \item The navigation information that guides the target vehicle toward the destination. We densely spread a set of navigation points in the route of target vehicle and include the relative position as observation. The target vehicle is not required to hit the navigation points but instead overpasses them. 
    \item The surrounding information including the relative positions and headings of the closest 4 traffic vehicles.
\end{itemize}

\textbf{Reward Scheme.} The reward function is composed of four parts as follows:
\begin{equation}
\label{eq:reward-functgion}
    R = c_{1}R_{disp} + c_{2}R_{speed} + c_{3}R_{steering} + c_{4}R_{term}.
\end{equation}
The \textit{displacement reward} $R_{disp} = d_t - d_{t-1}$, wherein the $d_t$ and $d_{t-1}$ denotes the longitudinal coordinates of the target vehicle in the current lane of two consecutive time steps, provides dense reward to encourage agent to move forward. 
The \textit{speed reward} $R_{speed} = v_t/v_{max}$ incentives agent to drive fast. $v_{t}$ and $v_{max}$ denote the current velocity and the maximum velocity ($120\ km/h$), respectively.
To prevent agent from swaying the vehicle left and right, we penalize it for large changes of steering by imposing a \textit{steering penalty} $R_{steering} = -|a_{1,t} - a_{1,t-1}| \cdot  v_{t}/v_{max}$, $a_{1,t}$ denotes the input steering at $t$.
We also define a sparse \textit{terminal reward} $R_{term}$, which is non-zero only at the last time step. At that step, we set $R_{disp} = R_{speed} = R_{steering}=0$ and assign $R_{term}$ according to the terminal state.
$R_{term}$ is set to $+20$ if the vehicle reaches the destination, $-10$ for crashing others, and $-5$ for violating the traffic rule.
We set $c_1 = c_4 = 1$ and $c_2 = c_3 = 0.1$. Sophisticated reward engineering may provide a better reward function, which we leave for future work.

\textbf{Split of training set and test set.}
PGDrive can produce a deterministic scene for each fixed random seed. This allows us to split the generated scenes into two sets: the training set and test set, through using exclusive random seeds.

\textbf{Evaluation Metrics.} We evaluate a given driving agent for multiple episodes and define the ratio of episodes where the agent arrives the destination as \textit{success rate}. The definition is the same for \textit{traffic rule violation rate} (namely driving out of the road) and the \textit{crash rate} (crashing other vehicles).
Compared to episodic reward, the success rate is a more suitable measurement when evaluating generalization, because we have a large number of scenes with different properties such as the road length and the traffic density, which leads the reward varying drastically across different scenes.

We fix the lane width to 3.5m, the number of lanes to 3 and traffic density to 0.1 by default. All experiments are repeated five times and we report the average results. 

\subsection{Results on Generalization}
\label{sect:result-on-generalization}
The overfitting happens if an agent achieves high performance in the training environments, but performs poorly in test environments that it has never been trained on.
On the contrary, a learned agent is considered to have good generalization if it works well on unseen environments and has small performance gap between training and test sets. 
We conduct a series of experiments to evaluate the generalization and investigate the improvement brought by procedural generation.
We create a suite of 7 environment sets varying the number of included training scenes ($N$) from 1 to 1000 and train policies on each set separately and then evaluate them on the same test set.
We train the agents with two popular RL algorithms respectively, PPO~\cite{schulman2017proximal} and SAC~\cite{haarnoja2018soft}, based on the implementation in RLLib~\cite{liang2018rllib}.
As shown in Fig.~\ref{fig:main-result}, the result of improved generalization is observed in the agents trained from both RL algorithms: First, the overfitting happens if the agent is not trained with a sufficiently large training set. 
When $N = 1$ where the agent is trained in a single map, we can clearly see the significant performance gap between when run learned policies in the training set and test set. 
Second, the generalization ability of agents can be greatly improved if the agents are trained in more environments. As the number of training scenes $N$ increases, the final test performance keeps increasing. 
The overfitting is alleviated and the test performance can match the training performance when $N$ is higher. SAC shows better generalization compared to PPO. 
The experimental results clearly show that increasing the diversity of training environments can significantly increase the generalization of RL agents, which validates the strength of the procedural generation introduced in PGDrive.

\subsection{Safety Generalization}
We define a new suite of environments for benchmarking the \textit{safety generalization ability} of end-to-end driving. 
As shown in Fig.~\ref{fig:maps}C, we randomly display static and movable obstacles in the traffic. We do not terminate the episode if the target vehicle crashes with those objects as well as traffic vehicles, as in Sec.~\ref{sect:result-on-generalization}. 
Instead, we allow agent to continue driving but record the crash with a cost $+1$. So the driving task is further formulated as a constrained MDP~\cite{achiam2017constrained,safety_gym_Ray2019}. 
We test the \textit{reward shaping method}, which assigns the crash penalty as negative reward following the reward scheme in Sec.~\ref{sect:exp-setting}, as well as the \textit{Lagrangian method}~\cite{safety_gym_Ray2019}:
\begin{equation}
    \max_{\theta} \min_{\lambda \ge 0} = \mathop{E}_{\tau}[R_{\theta}(\tau) - \lambda( C_{\theta}(\tau)-d)],
    \label{eq:lagrangian}
\end{equation}
wherein $R_{\theta}(\tau)$ and $C_{\theta}(\tau)$ are the reward objective and the cost objective respectively and $\theta$ is the policy parameters. $d$ is a given cost threshold, which is set to $3$. 
We use PPO for both methods.
For Lagrangian method, an extra cost critic network is used to approximate the cost advantage to compute the cost objective in Eq.~\ref{eq:lagrangian}. 
The Lagrangian multiplier $\lambda$ is independently updated before updating the policy at each training iteration.

\begin{figure}[!t]
    \centering
    \includegraphics[width=0.9\linewidth]{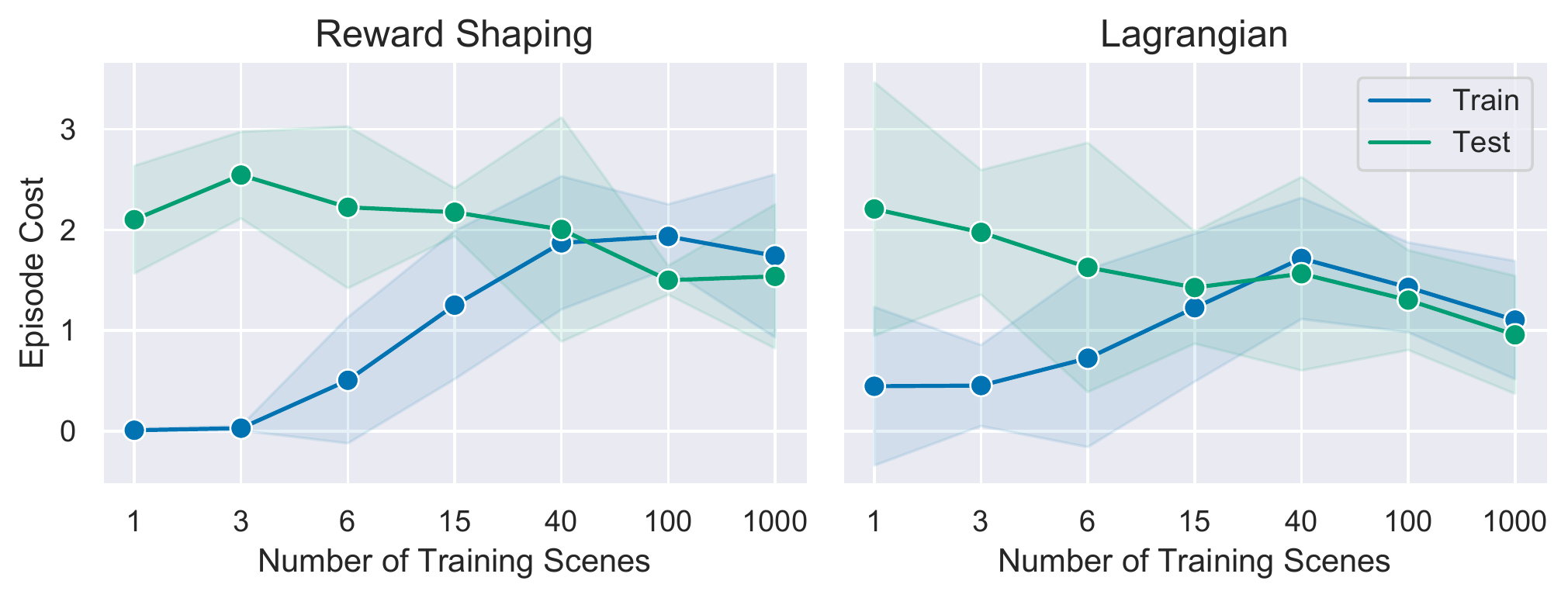}
    \caption{The episode cost of the trained policies in safety generalization suite. Both methods demonstrate overfitting and have poor safety performance in test time if trained with few training scenes. 
    }
    \label{fig:safety-result}
\end{figure}

We train both methods under different number of training scenes and demonstrate the training and test episode cost in Fig.~\ref{fig:safety-result}. We observe that the success rate follows the same tendency in Fig.~\ref{fig:main-result}, therefore we only present the plot of cost versus the number of training scenes here.
Both methods achieve high test cost even the training cost is low when training with few environments and improve the safety generalization when training diversity increases. Lagrangian method outperforms vanilla reward shaping method and reduces the test cost by around $50\%$.
This experimental result reveals the overfitting of safety, which is a critical research topic if we want to apply the end-to-end driving system to the real-world.

\begin{figure}[!t]
    \centering
    \hfill%
    \subfigure[\hspace{-2em}]{
        \includegraphics[width=0.46\linewidth]{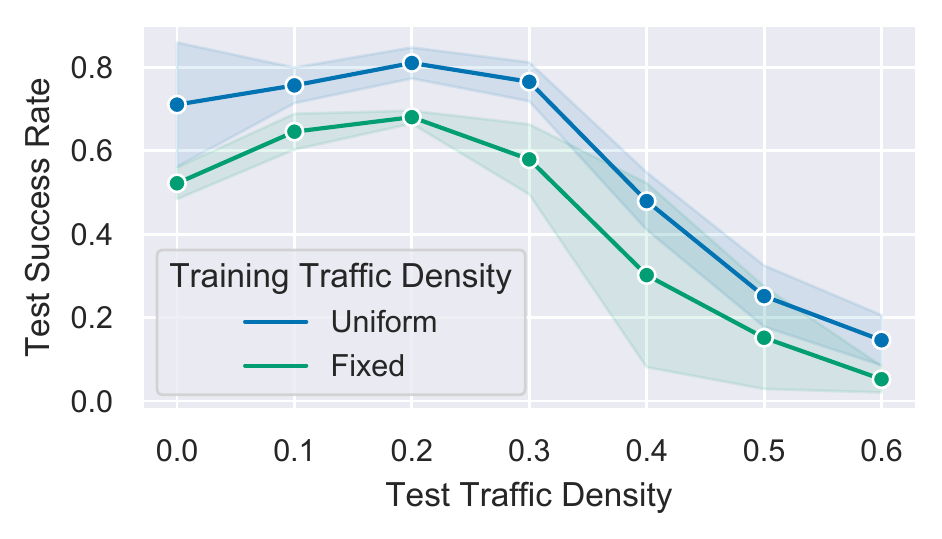}
    \label{fig:change-density}
    }    \hfill%
    \subfigure[\hspace{-2em}]{
        \includegraphics[width=0.46\linewidth]{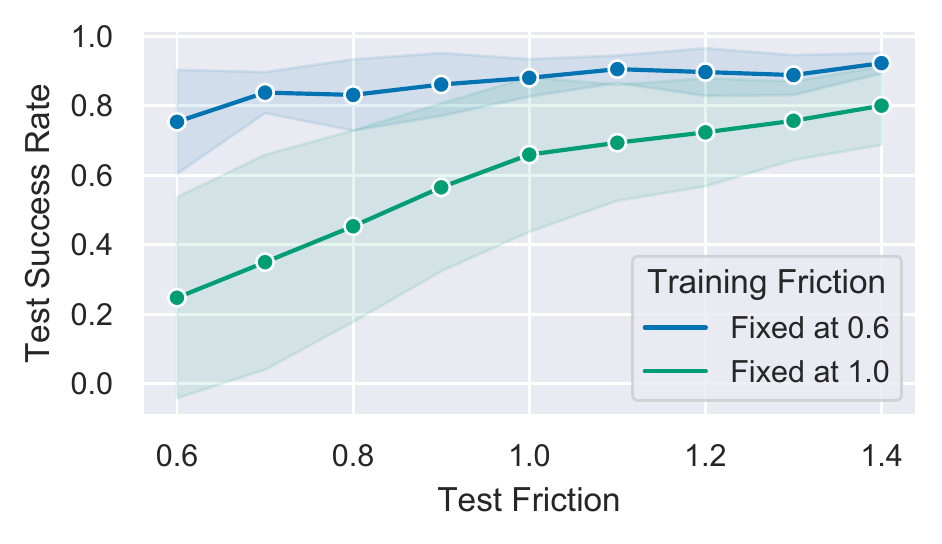}
    \label{fig:change-friction}
    }    \hfill%
    \vspace{-0.5em}
    \caption{
    (a) Agents demonstrate overfitting to the traffic density. The ``Fixed'' agent means the density is set to 0.1, while the ``Uniform'' agent varying the traffic density from 0 to 0.4 during training. 
    (b) Agents trained with wheel friction coefficient 0.6 have better generalization compared to those with 1.0 friction coefficient.
}
\end{figure}

\subsection{Factors Relevant to Generalization}
\label{sect:factors-relevant-to-generalization}
To better understand the contribution of the procedural generation to the improved generalization, we investigate two configurable factors of the environment that affect the generalization result.
As shown in Fig.~\ref{fig:change-density}, we find that randomizing the traffic density can improve the generalization of agents under the ``unseen traffic flow''. The agent trained with varying traffic density in range $[0.0, 0.4]$ consistently outperforms the agent trained in the density 0.1, even in the environment that both of them have not encountered during training, when the traffic density is set to 0.5 and 0.6.
Besides, the friction coefficient is also a critical factor that influence generalization, as shown in Fig.~\ref{fig:change-friction}.

\begin{figure}[!t]
    \centering\hfill
    \begin{minipage}{0.425\linewidth}
    \centering
    \includegraphics[width=\linewidth]{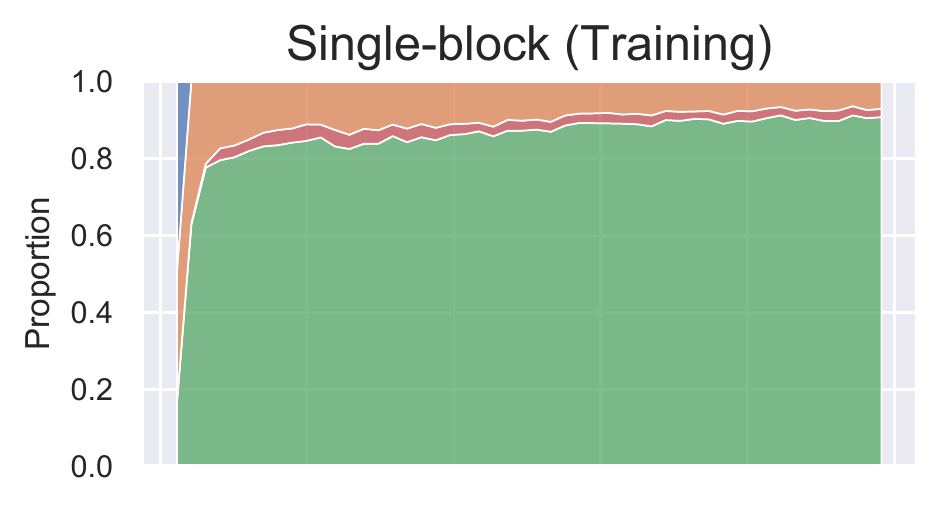}
    \end{minipage}\hfill
    \begin{minipage}{0.425\linewidth}
    \centering
    \includegraphics[width=\linewidth]{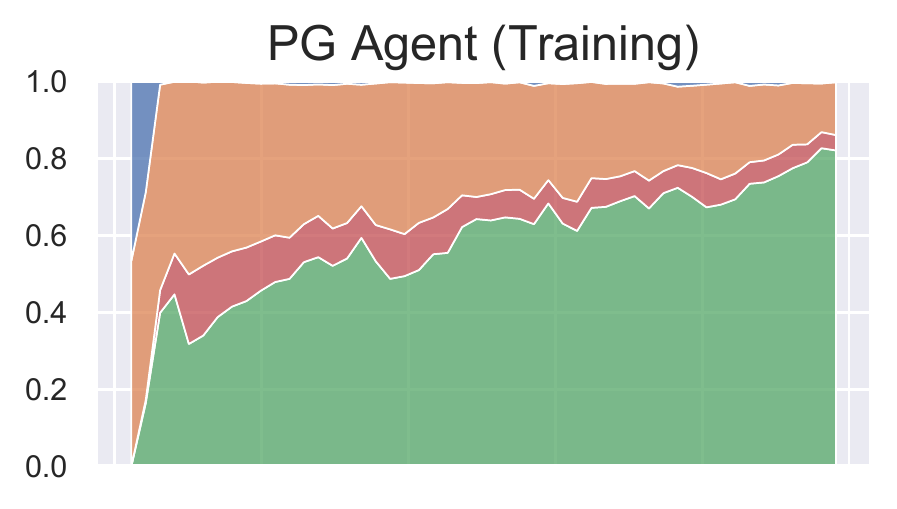}
    \end{minipage}\hfill ~\\
    \centering\hfill
    \begin{minipage}{0.425\linewidth}
    \centering
    \includegraphics[width=\linewidth]{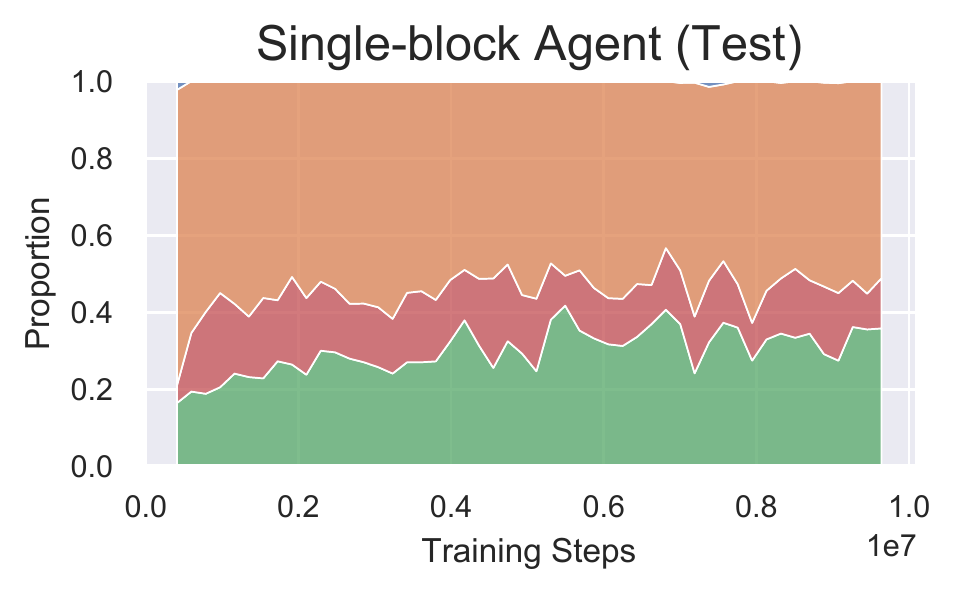}
    \end{minipage}\hfill
    \begin{minipage}{0.425\linewidth}
    \centering
    \includegraphics[width=\linewidth]{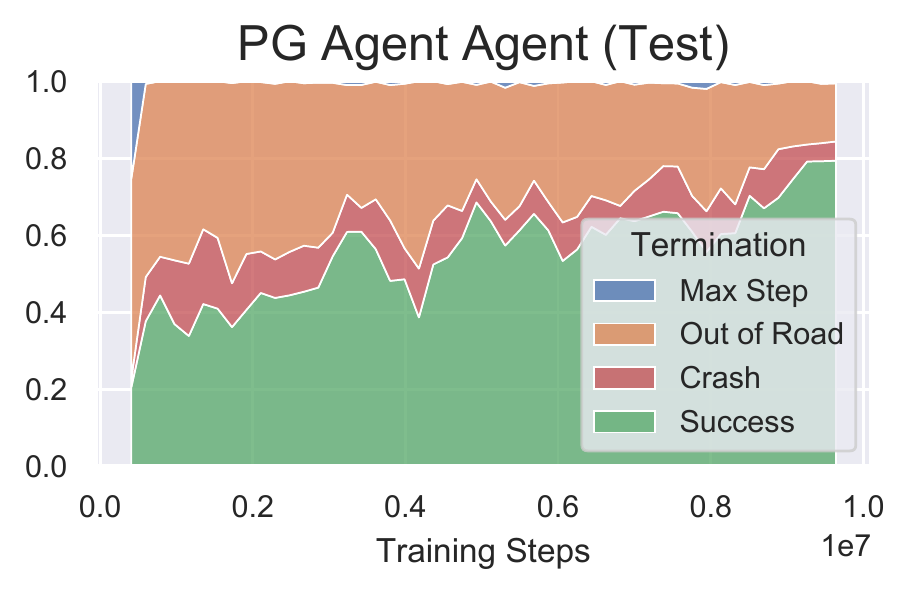}
    \end{minipage}\hfill ~\\
    \caption{Compared to the agents trained in multi-blocks environments (called PG Agent), agents trained in single-block environments can not generalize to complex environments.}
    \label{fig:PG-useful}
\end{figure}

We further conduct an experiment to show that an agent specialized on solving all types of blocks separately can not solve a complex driving map composed of multiple block types.
We compare two agents: 1) \textit{PG Agent} trained in 100 environments where each environment has 3 blocks, and 2) \textit{Single-block Agent} trained in 300 environments where each environment contains only 1 block. We evaluate them in the same test set of the environments with 3 blocks in each.
Fig.~\ref{fig:PG-useful} shows that both agents can solve their training tasks, but they show different test performance.
Agent trained on maps generated by PG performs better than agents trained on separate blocks. The results indicate that training agents in separate scenarios can not lead to good performance in complex scenarios. The procedurally generated maps which contain multiple blocks types lead to better generalization of the end-to-end driving.

\section{CONCLUSION}

We introduce the PGDrive, an open-ended and highly customizable driving simulator with the feature of procedural generation. Multiple road blocks with configurable settings are first defined and then assembled into diverse driving environments by the proposed Block Incremental Generation algorithm. The experimental results show that increasing the diversity of training environments can substantially improve the generalization of the end-to-end driving. 




\bibliography{cite}
\bibliographystyle{IEEEtran}

\section*{APPENDIX}
\subsection{Abstraction in PGDrive}

\begin{figure}[!h]
    \centering
   	\includegraphics[width=0.35\linewidth]{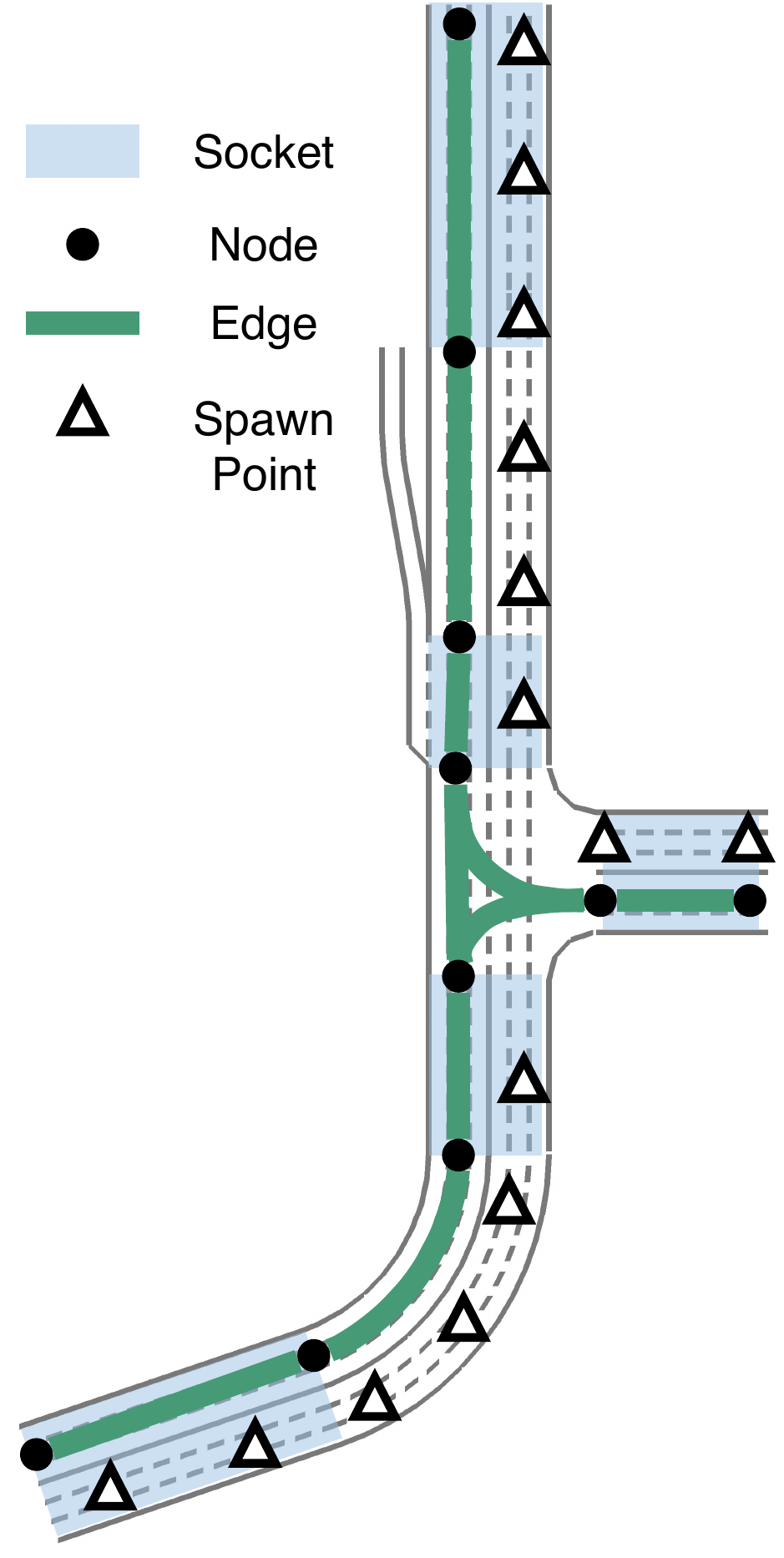}
    \caption{Abstraction of road network mentioned in Sec.~\ref{sect:abstraction-of-scene}.}
    \label{fig:node-edge}
\end{figure}

We present detailed structure of a simple 3-block road network in Fig.~\ref{fig:node-edge}. The spawn points are uniformly scattered in the road. The sockets are a straight road shared by two interconnected blocks.

\subsection{Learning Curves}

\begin{figure*}[!h]
    \centering
    \begin{minipage}{\linewidth}
    \includegraphics[width=\linewidth]{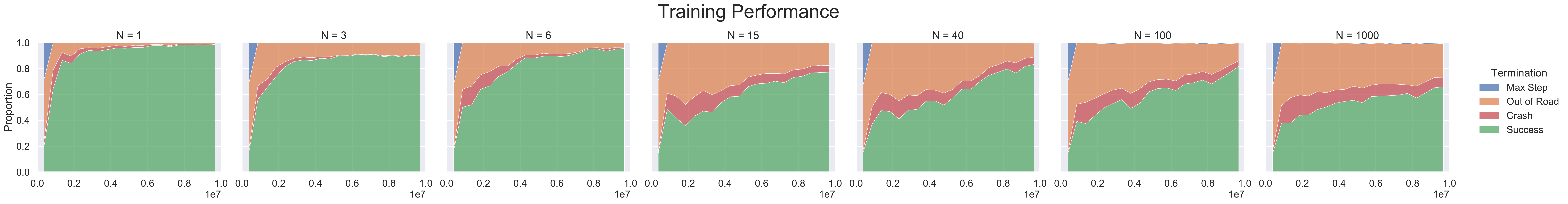}
    \end{minipage}~\\
    \vspace{0.5em}
    \begin{minipage}{\linewidth}
    \includegraphics[width=\linewidth]{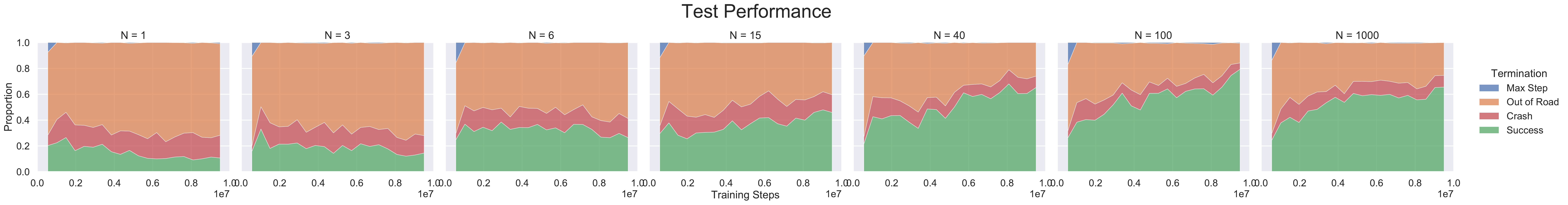}
    \end{minipage}
    \caption{The generalization result of the agents trained with on-policy algorithm PPO. Increasing the diversity of the training set substantially improves the generalization ability of end-to-end driving agents. The first row shows the training performance and the second row shows the test performance over training steps. The test set is the same set of held-out maps. Each column presents the performance of an agent trained on $N$ training maps. Increasing $N$ leads to the improvement of test success rate, which indicates the agent's generalization is improved.}
    \label{fig:main-result-full}
\end{figure*}

\begin{figure*}[!h]
    \centering
    \begin{minipage}{\linewidth}
    \includegraphics[width=\linewidth]{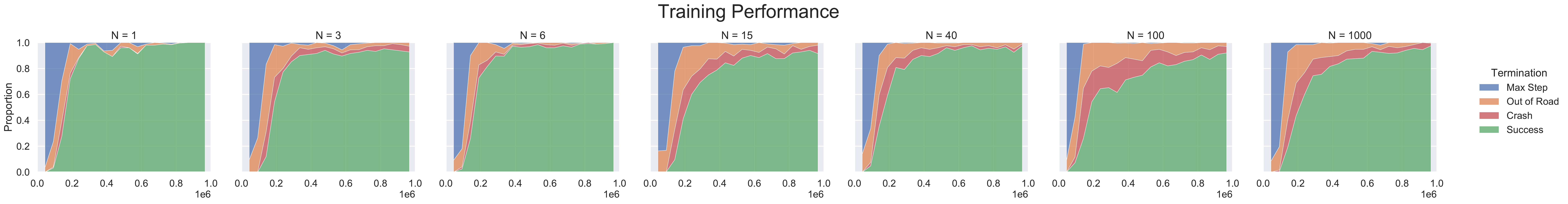}
    \end{minipage}~\\
    \vspace{0.5em}
    \begin{minipage}{\linewidth}
    \includegraphics[width=\linewidth]{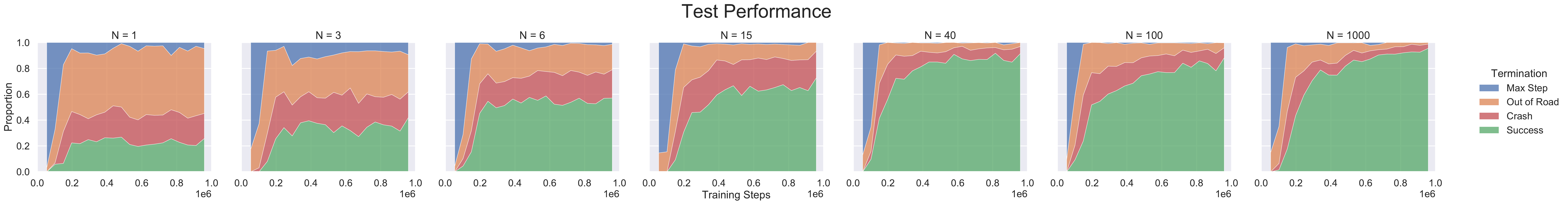}
    \end{minipage}
    \caption{We repeat the generalization experiments for the agents trained with the off-policy algorithm SAC. The result is similar to the one of the agents trained with PPO algorithm, but SAC algorithm brings more stable training performance.}
    \label{fig:main-result-sac-full}
\end{figure*}

Fig.~\ref{fig:main-result-full} and Fig.~\ref{fig:main-result-sac-full} present the detailed learning curves of the generalization experiments. Both methods show decreasing training performance and increasing test performance when the training diversity boosts. However, SAC converges much faster and presents more stable performance compared to PPO.

\begin{figure*}[!h]
    \centering
    \includegraphics[width=\linewidth]{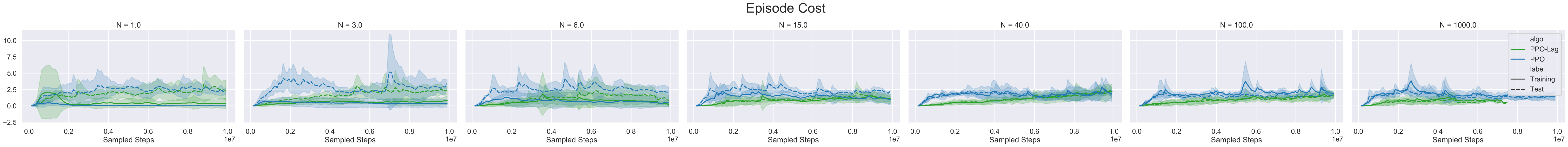}
    \caption{The episode cost of two methods in training and test scenarios. For both methods, the gaps of training performance and test performance get closer when increasing the training environments.}
    \label{fig:safety-full}
\end{figure*}

Fig.~\ref{fig:safety-full} present the detailed episode cost of both reward shaping method and Lagrangian method in training and test scenarios. When training with few environment, the test episode cost is large 
This demonstrates that the safety performance of policies is also over-fitted, as the prime driving performance, when training with

\subsection{Hyper-parameters}
We present the hyper-parameters for both PPO and SAC in this section.

\begin{table}[!h]
\centering
\centering
\caption{Hyper-parameters of PPO}
\begin{tabular}{@{}ll@{}}
\toprule
Hyper-parameter             & Value  \\ \midrule
$\lambda$                   & 1.0 \\
Discount Factor ($\gamma$)                    & 0.99 \\
Number of SGD epochs    & 20     \\
Learning Rate               & 0.00005 \\
Maximum Sampled Steps  & $1\times 10^7$ \\
SGD Minibatch Size & 100 \\
Training Batch Size & 30,000 \\
Number of Parallel Workers & 10 \\
Sampled Batch Size per Worker & 200 \\
Number of Random Seeds                & 5 \\ \bottomrule
\end{tabular}
\end{table}
\begin{table}[!h]
\centering
\caption{Hyper-parameters of SAC}
\begin{tabular}{@{}ll@{}}
\toprule
Hyper-parameter             & Value  \\ \midrule
Training Batch Size    & 256     \\
Maximum Steps & $1\times 10^6$ \\
Steps that Learning Starts & 10000 \\
Learning Rate               & 0.0001 \\
Target Update Coefficient ($\tau$)   & 0.005  \\
Prioritized Replay & True \\
Discount Factor ($\gamma$)                    & 0.99 \\
Number of Random Seeds                & 5 \\ \bottomrule
\end{tabular}
\end{table}

\subsection{Driving Scenes of Different Traffic Density}

\begin{figure*}[!h]
    \centering
   	\includegraphics[width=\linewidth]{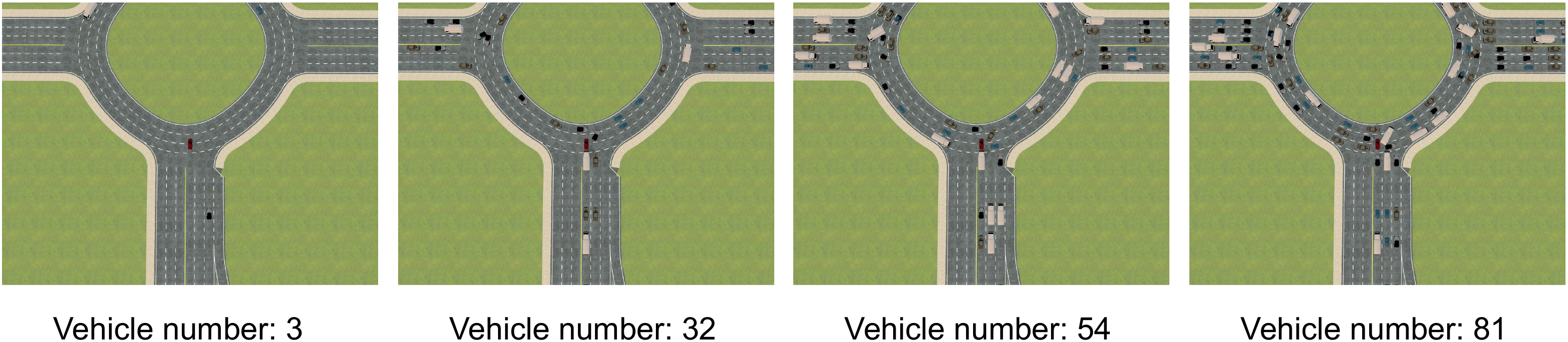}
    \caption{Driving scenes with different traffic density. From left to right, the traffic density of the scene is set to 0.1, 0.3, 0.5, 1.0, respectively.}
    \label{fig:varying-traffic-density}
\end{figure*}

In Sec.~\ref{sect:factors-relevant-to-generalization}, we demonstrate the results varying the training-time traffic density. In Fig.~\ref{fig:varying-traffic-density}, we showcase the driving scene with same road network and different traffic density.

\end{document}